\documentclass[10pt, conference, compsocconf]{IEEEtran}
\IEEEoverridecommandlockouts
\usepackage{cite}
\usepackage{amsmath,amssymb,amsfonts}
\usepackage{algorithmic}
\usepackage{graphicx}
\usepackage{textcomp}
\usepackage{xcolor}
\usepackage{colortbl}
\def\BibTeX{{\rm B\kern-.05em{\sc i\kern-.025em b}\kern-.08em
    T\kern-.1667em\lower.7ex\hbox{E}\kern-.125emX}}
\usepackage{booktabs}
\usepackage[caption=false]{subfig}
\usepackage{svg}
\usepackage{url}
\usepackage{multicol}
\usepackage{multirow}
\usepackage{color}
\usepackage{hhline}
\usepackage{pifont}
\usepackage{comment}
\usepackage{cleveref}
\usepackage{tablefootnote}
\usepackage[greek,english]{babel}
\usepackage{flushend}
\usepackage[ruled,vlined]{algorithm2e}
\definecolor{mynicegreen}{RGB}{11,102,35}
\usepackage[group-separator={,}]{siunitx}

\newcommand{\model}{{\textsc{SEAL}}}

 \newcommand{\squishlist}{
	\begin{list}{$\bullet$}
		{ \setlength{\itemsep}{0pt}
			\setlength{\parsep}{3pt}
			\setlength{\topsep}{3pt}
			\setlength{\partopsep}{0pt}
			\setlength{\leftmargin}{1.5em}
			\setlength{\labelwidth}{1em}
			\setlength{\labelsep}{0.5em} } }
	
	\newcommand{\squishlisttwo}{
		\begin{list}{$\bullet$}
			{ \setlength{\itemsep}{0pt}
				\setlength{\parsep}{0pt}
				\setlength{\topsep}{0pt}
				\setlength{\partopsep}{0pt}
				\setlength{\leftmargin}{2em}
				\setlength{\labelwidth}{1.5em}
				\setlength{\labelsep}{0.5em} } }
		
		\newcommand{\squishend}{
	\end{list}}

\begin{document}

\title{Semantically Encoding Activity Labels for Context-Aware Human Activity Recognition}

\author{\IEEEauthorblockN{%
Wen Ge\textsuperscript{\textsection},
Guanyi Mou\textsuperscript{\textsection},
Emmanuel O. Agu,
Kyumin Lee}
\IEEEauthorblockA{\textit{Computer Science Department}, 
\textit{Worcester Polytechnic Institute}
Worcester, MA, USA \\
{\{wge, gmou, emmanuel, kmlee\}}@wpi.edu}
}
\maketitle
\begingroup\renewcommand\thefootnote{\textsection}
\footnotetext{Equal contribution}
\endgroup


\begin{abstract}
Prior work has primarily formulated Context-Aware Human Activity Recognition (CA-HAR) as a multi-label classification problem, where model inputs are time-series sensor data and target labels are binary encodings representing whether a given activity or context occurs. These CA-HAR methods either predicted each label independently or manually imposed  relationships using graphs. However, both strategies often neglect an essential aspect: activity labels have rich semantic relationships. For instance, walking, jogging, and running activities share similar movement patterns but differ in pace and intensity, indicating that they are semantically related. Consequently, prior CA-HAR methods often struggled to accurately capture these inherent and nuanced relationships, particularly on datasets with noisy labels typically used for CA-HAR or situations where the ideal sensor type is unavailable (e.g., recognizing speech without audio sensors).
To address this limitation, we propose Semantically Encoding Activity Labels (SEAL), which leverage Language Models (LMs) to encode CA-HAR activity labels to capture semantic relationships. LMs generate vector embeddings that preserve rich semantic information from natural language. Our {\model} approach encodes input-time series sensor data from smart devices and their associated activity and context labels (text) as vector embeddings. During training, {\model} aligns the sensor data representations with their corresponding activity/context label embeddings in a shared embedding space. At inference time, {\model} performs a similarity search, returning the CA-HAR label with the embedding representation closest to the input data. Although LMs have been widely explored in other domains, surprisingly, their potential in CA-HAR has been underexplored, making our approach a novel  contribution to the field.
Our {\model} approach has been rigorously evaluated on three real-world datasets, demonstrating its superior performance. It consistently outperforms state-of-the-art methods by 7.8\% to 22.6\% in MCC and 3.9\% to 8.4\% in Macro-F1. Furthermore, {\model} performance is agnostic to the data encoding framework utilized, enhancing performance by 4.7\% to 73.3\% in MCC and 2.6\% to 30.5\% in Macro-F1 across different data encoding models. This robust performance opens up new possibilities for integrating more advanced LMs into CA-HAR tasks. We share our code at {\url{https://github.com/GMouYes/SEAL}}.
\end{abstract}

\begin{IEEEkeywords}
Context-Aware Human Activity Recognition, Language Modeling, Semantic Encoding, Machine Learning
\end{IEEEkeywords}

\section{Introduction}
\begin{figure}
\vspace{-10pt}
    \centering
    \subfloat[Traditional approach.]{
    \includegraphics[width=0.35\linewidth]{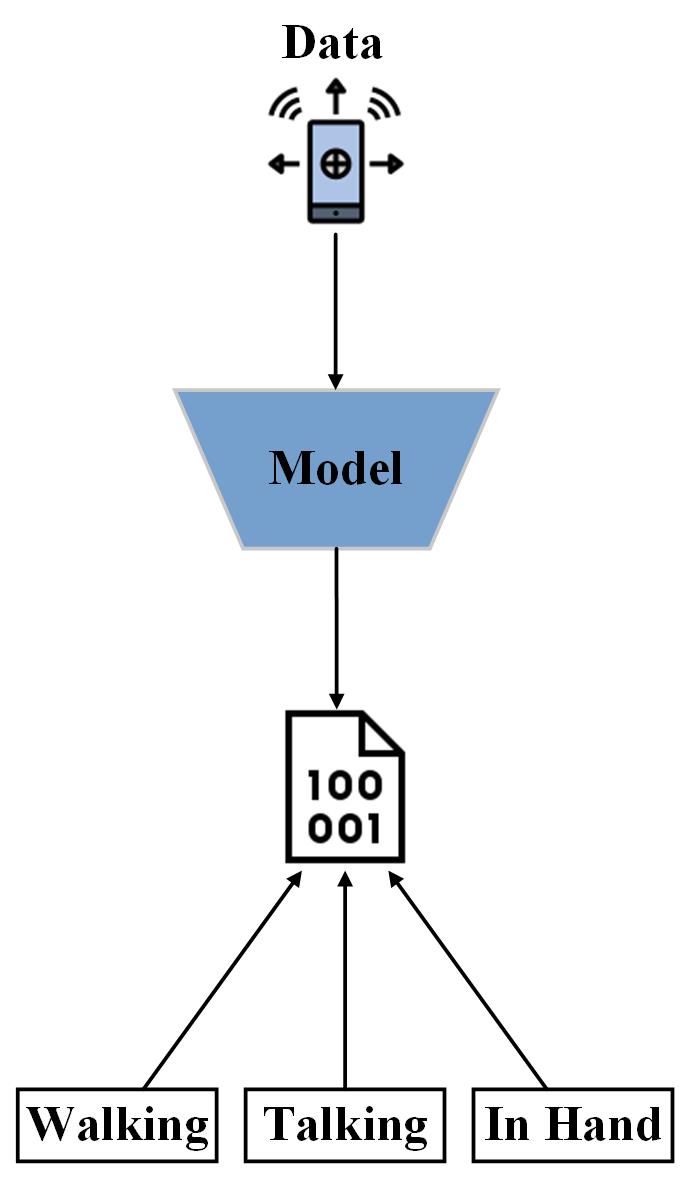}
    \label{fig:compareML}
    }
    \subfloat[Our approach.]{
    \includegraphics[width=0.35\linewidth]{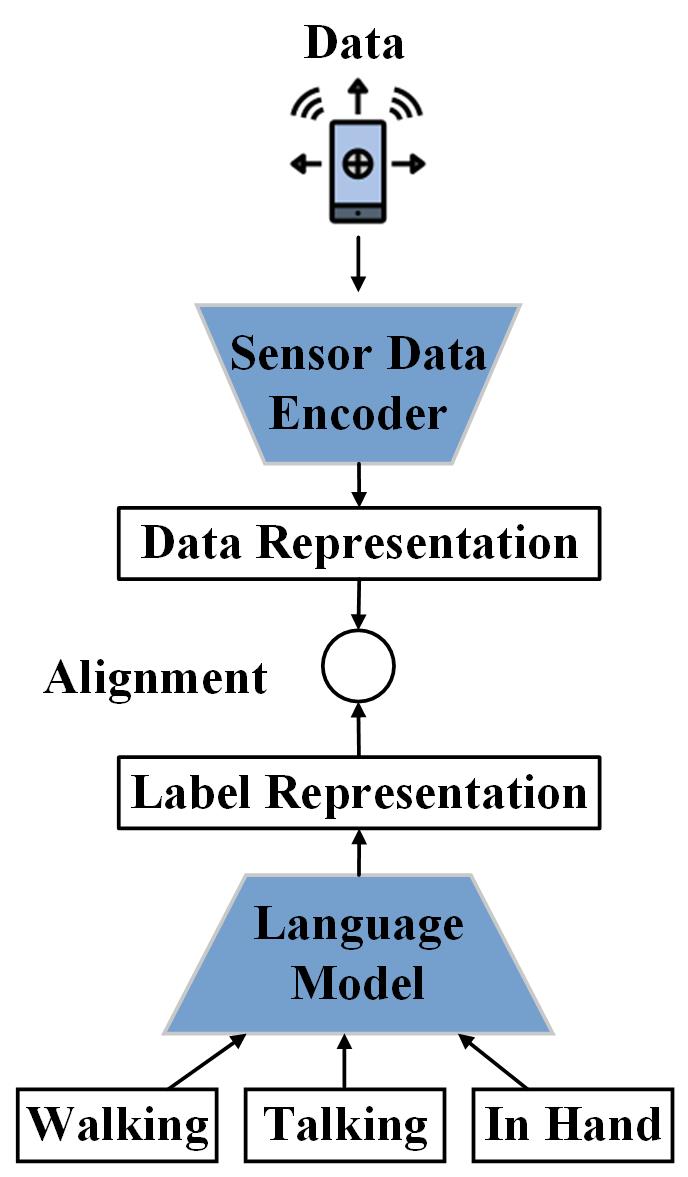}
    \label{fig:compareLM}
    }
    \caption{Comparison of traditional Machine Learning approach v.s. our multi-modality alignment approach. Traditional approaches directly map labels into binary values. In contrast, our approach leverages the language model to encode the semantic relationship between context and activities within high-dimensional vector representations and leads to better performance. }
    \label{fig:compareML_LM}
    \vspace{-10pt}
\end{figure}

Context-Aware Human Activity Recognition (CA-HAR) is a critical task that involves detecting human activities and their contexts using sensor data from devices such as smartphones and smartwatches~\cite{vaizman2017recognizing, bettini2020caviar}. 
Traditionally, CA-HAR has been formulated as a multi-label classification task, where the inputs are time-series sensor data, and the target labels are binary 0/1 encodings that indicate the absence/presence of specific activities and contexts. 
Previous research (Fig.~\ref{fig:compareML}) has focused on creating models that map sensor data to binary labels~\cite{vaizman2018context, ge2020cruft, ge2022qcruft}. More recent work has explored modeling label co-occurrences using manually defined, learned graphical representations to capture some relationships between labels~\cite{ge2023heterogeneous, ge2024deep, mondal2020new}. 

However, these approaches overlook a key consideration: activity labels have intrinsic, nuanced semantic relationships that are not captured when labels are assigned independent binary values, or missed by manually defined graphical relationships. For example, activities such as \textit{Walking}, \textit{Jogging}, and \textit{Running} share similar movement patterns but differ in pace and intensity, suggesting that they are semantically close. On the other hand,
misclassifying \textit{Trembling} as \textit{Standing} due to subtle signal changes could lead to missed early signs of medical conditions such as Parkinson's disease~\cite{de2022multiple}. Current models struggle to represent these subtle relationships, leading to suboptimal CA-HAR performance.

To bridge the gap, we propose a novel framework called Semantically Encoded Activity Labels ({\model}), a unique approach that leverages Language Models (LMs) \cite{devlin2018bert} and their associated rich embedding spaces to encode the rich semantics inherent in activity labels (Fig. \ref{fig:compareLM}). This novel approach is a departure from traditional methods and is likely to pique the interest of researchers and practitioners in the field. LMs have become foundational in Natural Language Processing (NLP) due to their ability to transform natural language into vector representations that preserve semantics. 

From a machine learning perspective, CA-HAR can be re-envisioned as an alignment task between two modalities: time-series data (sensor signals) and text (activity and context labels). During training, {\model} aligns the vector embeddings of sensor data with the embeddings of their corresponding activity/context labels. At inference time, it performs a similarity search to determine the label whose encoding is close to the input sensor data. By leveraging LMs and their associated embedding space with strong semantics, {\model} preserves semantic relationships between activity labels, capturing nuanced relationships between activities and contexts. Our work aligns with trending research in other domains~\cite{radford2021learning,jia2021scaling} where text-image alignments gained increasing momentum and enabled various real-world applications~\cite{betker2023improving,saharia2022photorealistic}.

\begin{figure}
\vspace{-10pt}
    \centering
    \subfloat[Other HAR models using Language Models (LMs).]{
    \includegraphics[width=.7\linewidth]{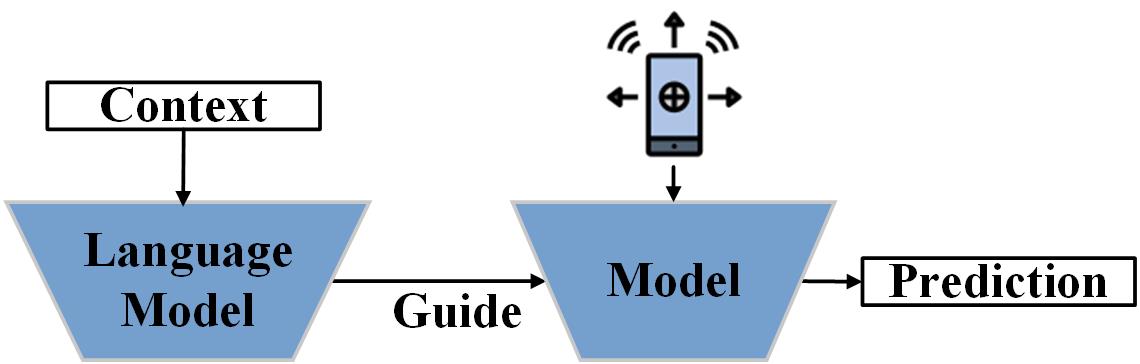}
    }
    \\
    \vspace{-5pt}
    \subfloat[Our design.]{
    \includegraphics[width=.7\linewidth]{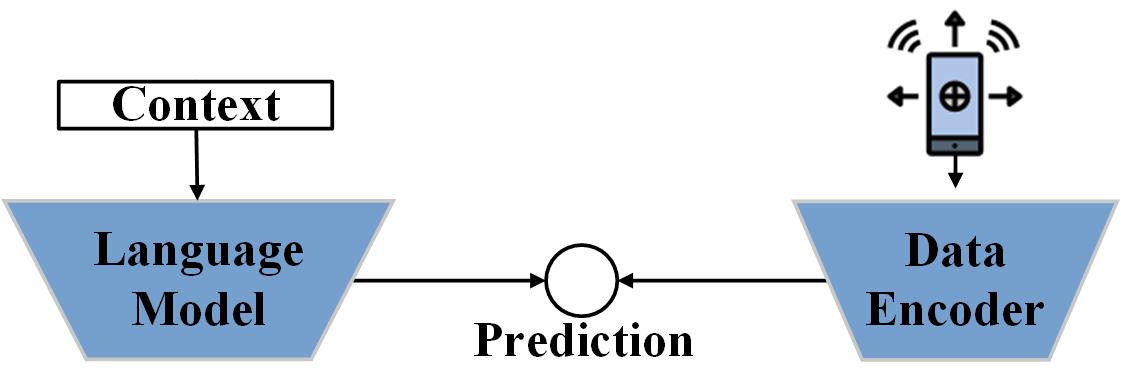}}
    \caption{Comparison of other HAR models using Language Models v.s. our design. While other approaches mainly use LMs as auxiliary components that provide guidance without directly participating in the decision-making process, our approach integrates LM as a primary contributor, allowing its active involvement in the activity recognition task.}
    \label{fig:otherlm}
    \vspace{-10pt}
\end{figure}

Although prior work has explored incorporating LMs into CA-HAR~\cite{zhou2023tent, arrotta2024contextgpt}, these efforts primarily leveraged the reasoning capabilities of LMs by using them as auxiliary modules that refine the outputs of CA-HAR models but did not integrate them as a core module in an end-to-end framework  (see Fig~\ref{fig:otherlm}). In contrast, our {\model} framework integrates LMs as a central component, serving as the primary encoder of activity labels. This innovative SEAL framework addresses the critical issue of semantic information loss inherent in binary label encodings, a significant limitation of prior CA-HAR methods. Furthermore, our {\model} approach achieves robust performance improvements and generalizes effectively, even on the relatively noisy datasets commonly used in CA-HAR. In this paper, we define context-aware human activity the as the $<$activity, phone placement$>$ tuple.

\smallskip In summary, our contributions are as follows: 
\squishlist 
\item We identify the loss of semantic relationship information caused by binary label encodings widely used in prior CA-HAR frameworks. We reframe CA-HAR as a time-series-to-text alignment task in a shared vector embedding space, a perspective that should enlighten our audience about the nuanced relationships between activity labels. 
\item We introduce {\model}, a novel framework that, for the first time, leverages language models to semantically encode activity labels and align these textual encodings with sensor data representations, improving activity recognition. 
\item We extensively evaluate {\model}, comparing it to five state-of-the-art baselines on three real-world CA-HAR datasets. Our detailed evaluations provide robust evidence of {\model}'s performance, demonstrating that it consistently outperforms baseline approaches, with MCC/Macro-F1 improvements ranging from 7.8\%/3.9\% to 22.6\%/8.4\%. This comprehensive evaluation should provide reassurance to our audience about the effectiveness of our approach. 
\item We provide rigorous analysis and statistical and visual evidence demonstrating our approach's effectiveness and contributions to the CA-HAR field. 
\squishend

\vspace{-5pt}
\section{Related Work}
\vspace{-5pt}
\subsection{Traditional HAR}

Early CA-HAR work trains separate machine learning models for each label independently~\cite{vaizman2017recognizing,gao2019human}, overlooking the complex relationships between context and activities. This approach proved inadequate as the number of activities encountered and recognized grew, leading to scalability issues and suboptimal performance. In recent years, deep learning techniques, such as MLP\cite{vaizman2018context}, CNN\cite{munzner2017cnn}, RNN\cite{mekruksavanich2021lstm,zhao2018deep}, and Hybrid  Networks\cite{ge2020cruft,ge2022qcruft}, have gained popularity in HAR as they implicitly model relationships between activity labels by training a unified model. Despite their success, these models lack explicit mechanisms to capture the relationships between context and activity labels, often resulting in a limited understanding of the context's influence on activity recognition.

\vspace{-3pt}
\subsection{Graph-based HAR}

To overcome the limitations of traditional HAR models, recent studies have delved into graph-based methods that explicitly encode relationships between context and activity labels. Martin \textit{et al.}~\cite{martin2018graph} proposed a GCN approach for HAR by modeling personalized mobility graphs from GPS trajectory. HAR-GCNN~\cite{mohamed2022har} leveraged chronological correlations between sequential activities to predict missing labels. HHGNN~\cite{ge2023heterogeneous} and DHC-HGL~\cite{ge2024deep} proposed modeling the activity and context labels as nodes in a heterogeneous hypergraph.
However, the manual definition of these graphs may restrict the expressiveness and adaptability of the learned representations.
Traditional and graph-based HAR works overlooked a crucial aspect of the problem: activity labels carry intrinsic semantic meanings. Employing independent and binary label encodings without capturing their semantics misses the nuanced, inherent relationships, similarities, and distinctions beyond simple co-occurrence and leads to sub-optimal performance. For example, while Walking, Jogging, and Running are semantically similar in movement patterns, they differ in pace and intensity. 

\subsection{Using Language Model for HAR}
Recent advancements in LMs have demonstrated their capabilities in understanding and processing natural language, inspiring many applications in various other domains, including computer vision~(CV)\cite{zhou2022learning} and health applications~\cite{yang2022large}. 
For instance, CLIP~\cite{radford2021learning} introduces an innovative method for learning visual representations using natural language supervision. Their model learns a multi-modal embedding space by jointly training an image and text encoder, optimizing for high cosine similarity between correct image-text pairs while minimizing it for incorrect pairs. 
Similarly, ALIGN~\cite{jia2021scaling} developed a scalable model that processes noisy image-alt-text pairs using a dual-encoder architecture, aligning image and text representations in a shared latent space through a contrastive loss. TS2ACT~\cite{xia2024ts2act} introduced a cross-modal contrastive learning framework for HAR, aligning time-series sensor data with web-sourced images. However, they relied on image-based augmentation rather than the inherent semantics of labels. Unlike SEAL, TS2ACT oversimplifies real-world scenarios by focusing on multi-class activity classification while ignoring co-occurring activities and contextual influences.
These developments have encouraged researchers to explore the potential of language models for human activity recognition (HAR) tasks. 
ContextGPT~\cite{arrotta2024contextgpt} translated high-level contextual information into natural language using prompt engineering and retrieved the most plausible activities from pre-trained Large Language Models (LLMs) with the given context. The retrieved information is then infused into a HAR model to enhance recognition performance. Their approach used LM as an \underline{auxiliary component} to help guide the model to focus more on plausible activities. Unlike their approach, {\model} integrates LM as a core module into an end-to-end model to capture the relationship between context and activities. It can be used directly for decision-making. Besides that, their reliance on LLMs for generating knowledge might lead to hallucinations and result in inconsistent activity predictions.

\begin{figure}
    \centering
    \vspace{-5pt}
    \includegraphics[width=.85\linewidth]{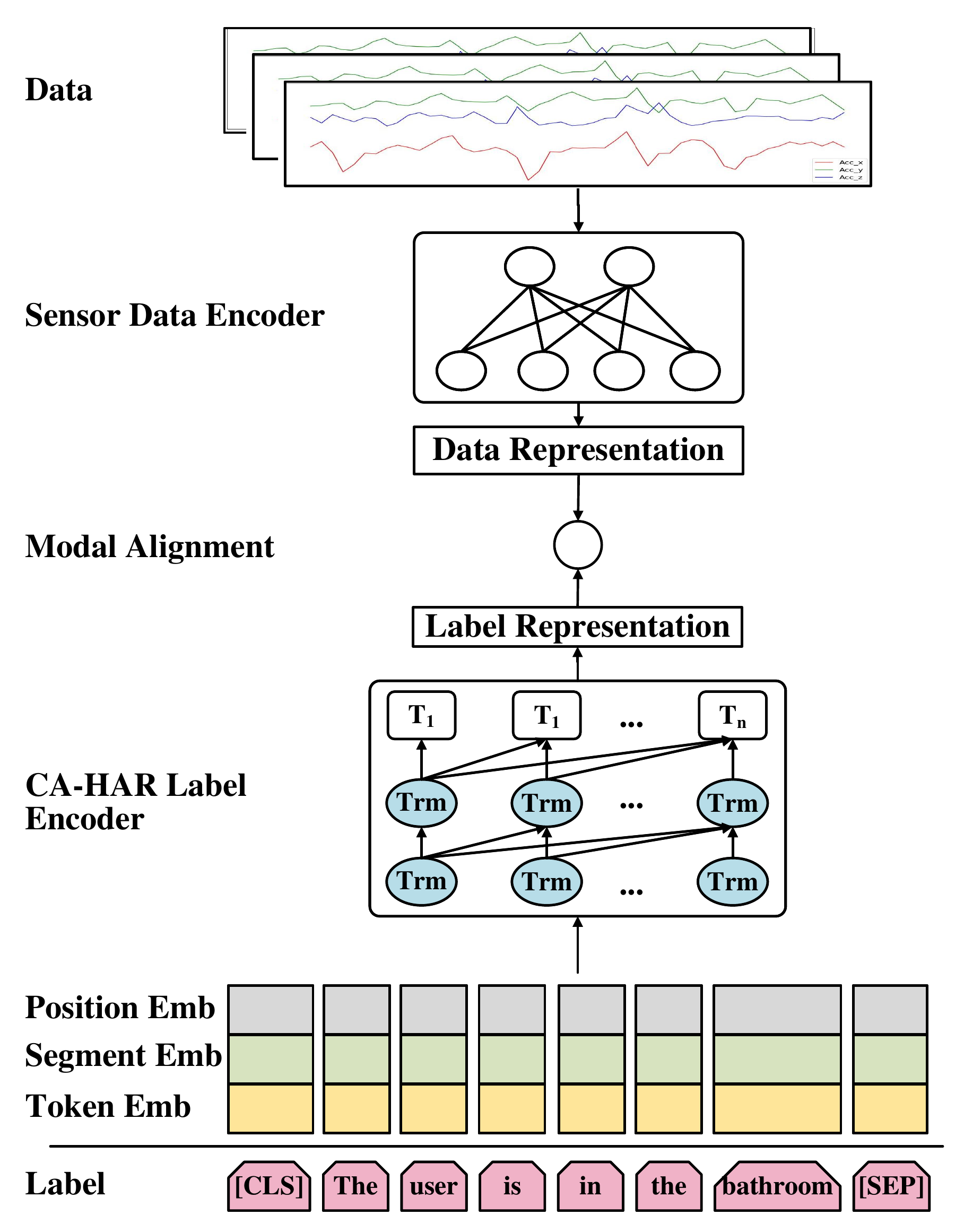}
    \vspace{-10pt}
    \caption{The {\model} framework consists of three main components: a Sensor Data Encoder, a CA-HAR Label Encoder, and a Modal Alignment. The Sensor Data Encoder transforms input sensor data into vector embedding representations, while the Label Encoder generates semantic label vector embedding representations from tokenized label sentences. Finally, the Modal Alignment component aligns the sensor data and CA-HAR label representations by maximizing their similarity, enabling {\model} to make accurate predictions. ``Trm'' are transformers modules within language models.}
    \label{fig:framework}
    \vspace{-10pt}
\end{figure}

\section{Proposed Framework}
\label{sec:framework}

\subsection{Overall Architecture}
\label{sec:overall}
Our primary goal is to align the time-series input with its corresponding textual labels while preserving the underlying semantic relationships and accounting for potential label co-occurrence. This formulation contrasts with traditional CA-HAR frameworks~\cite{vaizman2018context,ge2020cruft,ge2023heterogeneous,ge2024deep}, which use arbitrary binary encodings for labels and treat the task as a single-modal data-to-label mapping, $f: X \rightarrow Y_b$. In these approaches, $f$ is a learnable mapping function, and $Y_b$ represents the binary label format. However, binary encodings discard the semantic meanings of the original labels and fail to capture the inherent similarities, differences, and relationships between activities. In contrast, our approach uses an LM to encode the original text labels, retaining rich semantic relationships and ensuring that the labels' meaning preserved in the process. This enables a more accurate alignment between sensor data, activities and contexts. Figure~\ref{fig:framework} shows our architecture in detail, which consists of three sub-modules: 
\squishlist 
\item \textit{Sensor Data Encoder ($M_{data}$):} Encodes input time-series sensor data into vector embedding representations. 
\item \textit{CA-HAR Label Encoder ($M_{label}$):} Encodes the activity labels in textual form into vector embedding representations. 
\item \textit{Modal Alignment ($M_{align}$):} uses a custom objective function to aligns the vector embeddings generated by the sensor data and CA-HAR label encoders. 
\squishend 
Our key contributions lie mostly in the CA-HAR label encoder and modal alignment components. 

\begin{table*}[tbp]
\centering
\vspace{-10pt}
\caption{Average activity recognition performance on the \textit{WASH Scripted}, \textit{WASH Unscripted}, and \textit{Extrasensory} datasets, each cell contains MCC/Macro-F1 scores. Best Result: \textbf{Bolded}. Second Best Result: \underline{underlined}. }
\vspace{-5pt}
\scalebox{.85}{
\begin{tabular}{|l|c|c|c|c|c|c|c|c|r|}
\toprule
\textbf{Dataset} & \textbf{ExtraMLP} & \textbf{LightGBM} & \textbf{CRUFT} & \textbf{HHGNN} & \textbf{DHC-HGL} & \textbf{{\model}} & \textbf{Improv.(\%)} \\ \midrule
\textit{WASH Scripted} & 0.387 / 0.637 & \underline{0.411 / 0.657} & 0.409 / 0.652 & 0.389 / 0.640 & \underline{0.411 / 0.655} & \textbf{0.504 / 0.712} & \cellcolor{gray!30}22.6 / 8.4 \\ 
\midrule
\textit{WASH Uncripted} & 0.550 / 0.735 & 0.667 / 0.810 & 0.557 / 0.742 & 0.693 / 0.827 & \underline{0.808 / 0.897} & \textbf{0.921 / 0.959} & \cellcolor{gray!30}14.0 / \cellcolor{gray!30}6.9 \\\midrule
\textit{Extrasensory} & {0.636 / 0.787} & {0.710 / 0.837} & {0.769 / 0.871} & {0.808 / 0.896} & \underline{0.855 / 0.923} & \textbf{0.922 / 0.959} & \cellcolor{gray!30}{7.8 / \cellcolor{gray!30}3.9} \\
\bottomrule
\end{tabular}}
\label{tab:activity}
\vspace{-5pt}
\end{table*}

\begin{table}[ht]
\centering
\vspace{-10pt}
\caption{Examples of label transformation for language model input. For short-term actions, the transformed sentences explicitly included 'now' to emphasize the action as a transitional state occurring in the present moment.}
\vspace{-5pt}
\scalebox{.85}{
\begin{tabular}{|l|l|}
\toprule
\textbf{Original Label} & \textbf{Transformed Sentence} \\ \midrule
In Pocket & The user has a phone in their pocket. \\ \midrule
Bathroom & The user is in the bathroom .\\ \midrule
Talking on Phone & The user is talking on the phone. \\ \midrule
Sitting down (action) & The user is sitting down now. \\ \bottomrule
\end{tabular}}
\label{tab:sentence}
\vspace{-10pt}
\end{table}

\subsection{Sensor Data Encoder}
\label{sec:dataEncoder}
The sensor data encoder ($M_{data}$) transforms input time-series sensor data into vector representations. Existing methods follow two main approaches: 

\smallskip\noindent \underline{Automatically learn} feature representations from raw time series data using a CNN and/or Bi-LSTM as in~\cite{mekruksavanich2021lstm,vaswani2017attention}: $ \scriptsize M_{data}(X) = V_{data} \in \mathbb{R}^{n \times h_1}$.

\smallskip\noindent \underline{Extract handcrafted features}, followed by machine learning models to obtain refined feature representations as in~\cite{vaizman2018context}: 
\begin{equation}
\scriptsize
    M_{data}(T(X)) = V_{data} \in \mathbb{R}^{n \times h_1},  T:\mathbb{R}^{n \times s \times d} \rightarrow \mathbb{R}^{n \times h'} 
    \vspace{-5pt}
\end{equation}
where $h'$ and $h_1$ are hidden dimensions. 
%
%
While we utilized a two-layer MLP with activation and dropout to derive data representation from handcrafted features,
it is instructive to note that our framework is highly adaptable to other data encoding methods. In future, with minimal modification, researchers can explore additional modalities such as audio, image, and video, as well as alternative data encoders~\cite{li2019bi, vaswani2017attention} to further improve model performance. As demonstrated in Section~\ref{sec:analysis}, our {\model} framework imposes minimal restrictions on the data encoder and improves performance across different backbone models (e.g., MLPs~\cite{vaizman2018context}, RNNs~\cite{schuster1997bidirectional}, and GNNs~\cite{ge2024deep}).

\subsection{CA-HAR Label Encoder}
\label{sec:labelEncoder}
Unlike time-series input data, the CA-HAR labels are represented in textual form. Inspired by recent advances~\cite{radford2021learning}, we employ LMs to encode the labels into vector representations that preserve rich CA-HAR label semantics and relationships.
\begin{equation}
\scriptsize
    M_{label}(Y_t) = V_{label}, Str^{n \times C} \rightarrow \mathbb{R}^{n \times C \times h_2}
\end{equation}
In this paper, we finetune a pre-trained BERT-based model to generate a CA-HAR label representation. 
Labels are first rewritten into complete sentences using templates as shown in Table~\ref{tab:sentence}, 
then tokenized and fed into BERT to extract representations from the classification head: 
\begin{equation}
\scriptsize
\begin{aligned}
    Y'_t &= Tokenize(Rewrite(Y_t))\\
    V_{label} &= BERT(Y'_t)<cls>
\end{aligned}
\label{eq:bert}
\end{equation}
where $<cls>$ refers to the classification head.
While we recognize the availability of larger and more powerful language models~\cite{raffel2020exploring,liu2019roberta,petroni2019language,team2023gemini,open2023chatgpt}, the novelty of our approach is that it is LM-agnostic. We focus on demonstrating that incorporating an LM as a CA-HAR label encoder is effective. Future work can explore larger models towards incremental performance gains or efficient models~\cite{sanh2019distilbert} to reduce computational costs, and facilitate deployment in real-world applications.

\subsection{Modal Alignment}
\label{sec:alignmentModule}

The modal alignment module aligns the data encoder and label encoder representations in a shared vector space, where similarity is measured using a dot product. 
\begin{equation}
\scriptsize
\begin{aligned} 
V'_{data} &= MLP_d(V_{data}) \in \mathbb{R}^{n \times h}\\
V'_{label} &= MLP_l(V_{label}) \in \mathbb{R}^{n \times C \times h} \\
\hat{Y}_b &= dot(V'_{data}, V'_{label}) \in \mathbb{R}^{n \times C}
\end{aligned}
\end{equation}
During \model~training, the modal alignment module aligns the data and corresponding label vector embedding encodings by maximizing their similarity. During inference time, the model returns the labels with the LM encoding that are similar (close) to a given data encoding w.r.t thresholds for the multi-label problem. To optimize this alignment, we employ two loss functions depending on the nature of the labels. For mutually exclusive context labels such as phone placements, the Cross-Entropy (CE) loss is utilized.
\begin{equation}
    \label{eq:ce}
    \scriptsize
    L_{ce} = -\frac{1}{N}\sum_{n=1}^N \sum_{c=1}^C y_{n,c}log(\frac{exp(\hat{y}_{n,c})}{\sum_{i=1}^Cexp(\hat{y}_{n,i})})
\end{equation}
For CA-HAR activities that may co-occur (e.g.  \textit{Walking} while \textit{Talking On Phone}), Binary Cross-Entropy (BCE) is applied:
\begin{equation}
    \label{eq:bce}
\resizebox{0.91\hsize}{!}{%
    $L_{bce} = -\frac{1}{N}\sum_{n=1}^N \sum_{c=1}^C \omega_{n,c}\Big[y_{n,c}\log(\hat{y}_{n,c}) + (1-y_{n,c})\log(1-\hat{y}_{n,c}) \Big]$%
}
\end{equation}
In equations \ref{eq:ce} and \ref{eq:bce}, $N$ is the number of instances, $C$ is the number of classes, and $y_{n,c}$ and $\hat{y}_{n,c}$ are the ground truth and predicted values, respectively.

\section{Evaluation}
\label{sec:exp}
We rigorously evaluated {\model}'s performance on three CA-HAR datasets, namely WASH Scripted, WASH Unscripted, and Extrasensory. For more details about our dataset and experiment settings, please refer to Sec.~\ref{sec:dataset} and Sec.~\ref{sec:expSetting}. Average activity recognition performance is presented in Table~\ref{tab:activity}, and detailed experimental results on \textit{WASH Scripted} and \textit{Extrasensory} are listed in Table~\ref{tab:scripted_vs_baseline} and Table~\ref{tab:extra_vs_baseline}. {\model} uses MLP for data encoder in this section, while we discuss different data encoders' impact on the model's performance in Sec.~\ref{sec:analysis}.

\begin{table*}[htbp]
\centering
\vspace{-5pt}
\caption{Activity recognition performance on the \textit{WASH Scripted} dataset, each cell contains MCC/Macro-F1 scores.}
\vspace{-5pt}
\scalebox{.85}{
\begin{tabular}{|l|c|c|c|c|c|c|c|c|r|}
\toprule
\textbf{Activity} & \textbf{ExtraMLP} & \textbf{LightGBM} & \textbf{CRUFT} & \textbf{HHGNN} & \textbf{DHC-HGL} & \textbf{{\model}} & \textbf{Improv.(\%)} \\ \midrule
\textit{Lying Down}
& 0.183 / 0.513 & 0.185 / 0.520 & 0.202 / 0.526 & 0.221 / 0.540 & \underline{0.227 / 0.544} & \textbf{0.260 / 0.563} & \cellcolor{gray!30}14.5 / 3.5 \\ 
\textit{Sitting}
& 0.530 / 0.718 & \underline{0.550 / 0.739} & 0.492 / 0.690 & 0.522 / 0.716 & 0.509 / 0.709 & \textbf{0.646 / 0.796} & \cellcolor{gray!30}17.5 / 7.7 \\ 
\textit{Walking}
& 0.885 / 0.942 & 0.869 / 0.934 & 0.871 / 0.935 & 0.874 / 0.937 & \underline{0.899 / 0.949} & \textbf{0.909 / 0.954} & \cellcolor{gray!30}1.1 / 0.5 \\ 
\textit{Sleeping} 
& 0.505 / 0.701 & \textbf{0.596 / 0.763} & 0.519 / 0.712 & 0.477 / 0.684 & 0.503 / 0.701 & \underline{0.570 / 0.745} & -4.4 / -2.4 \\ 
\textit{Talking On Phone} 
& 0.393 / 0.632 & 0.528 / 0.723 & \underline{0.601 / 0.768} & 0.493 / 0.698 & 0.406 / 0.640 & \textbf{0.616 / 0.778} & \cellcolor{gray!30}2.5 / 1.3 \\ 
\textit{Bathroom} 
& 0.316 / 0.576 & 0.348 / 0.598 & \underline{0.362 / 0.614} & 0.313 / 0.577 & 0.344 / 0.602 & \textbf{0.528 / 0.725} & \cellcolor{gray!30}45.9 / 18.1 \\ 
\textit{Standing} 
& 0.356 / 0.598 & 0.345 / 0.597 & 0.380 / 0.622 & 0.342 / 0.591 & \underline{0.401 / 0.642} & \textbf{0.454 / 0.669} & \cellcolor{gray!30}13.2 / 4.2 \\ 
\textit{Jogging} 
& 0.632 / 0.786 & \underline{0.826 / 0.907} & 0.671 / 0.812 & 0.653 / 0.799 & 0.649 / 0.797 & \textbf{0.837 / 0.913} & \cellcolor{gray!30}1.3 / 0.7 \\ 
\textit{Jumping} 
& 0.429 / 0.656 & 0.576 / 0.754 & 0.545 / 0.731 & 0.526 / 0.721 & \underline{0.614 / 0.779} & \textbf{0.707 / 0.837} & \cellcolor{gray!30}15.1 / 7.4 \\ 
\textit{Running} 
& \underline{0.715 / 0.840} & 0.713 / 0.841 & 0.592 / 0.761 & 0.683 / 0.821 & 0.637 / 0.791 & \textbf{0.808 / 0.896} & \cellcolor{gray!30}13.0 / 6.5 \\ 
\textit{Stairs-Going Down} 
& \underline{0.460 / 0.680} & 0.351 / 0.608 & 0.371 / 0.618 & 0.419 / 0.652 & 0.440 / 0.666 & \textbf{0.557 / 0.742} & \cellcolor{gray!30}21.1 / 9.1 \\
\textit{Stairs-Going Up} 
& 0.316 / 0.589 & 0.274 / 0.567 & 0.280 / 0.568 & 0.308 / 0.589 & \underline{0.377 / 0.636} & \textbf{0.408 / 0.644} & \cellcolor{gray!30}8.2 / 1.3 \\
\textit{Typing} 
& 0.612 / 0.772 & \textbf{0.770 / 0.873} & 0.744 / 0.859 & 0.558 / 0.737 & 0.572 / 0.747 & \underline{0.745 / 0.857} & -3.2 / -1.8 \\
\textit{Coughing} 
& 0.195 / 0.532 & 0.155 / 0.521 & 0.199 / 0.533 & 0.182 / 0.528 & \underline{0.250 / 0.560} & \textbf{0.300 / 0.589} & \cellcolor{gray!30}20.0 / 5.2 \\ 
\textit{Sneezing} 
& 0.170 / 0.520 & 0.153 / 0.522 & 0.200 / 0.536 & 0.181 / 0.527 & \underline{0.242 / 0.561} & \textbf{0.302 / 0.591} & \cellcolor{gray!30}24.8 / 5.3 \\
\textit{Trembling} 
& 0.415 / 0.647 & 0.384 / 0.632 & \underline{0.453 / 0.673} & 0.377 / 0.624 & 0.390 / 0.635 & \textbf{0.607 / 0.771} & \cellcolor{gray!30}34.0 / 14.6 \\
\textit{Laying Down (a)} 
& 0.190 / 0.524 & 0.186 / 0.526 & 0.223 / 0.540 & 0.202 / 0.531 & \underline{0.229 / 0.550} & \textbf{0.247 / 0.558} & \cellcolor{gray!30}7.9 / 1.5 \\ 
\textit{Sitting Down (a)} 
& 0.131 / 0.502 & 0.119 / 0.505 & 0.136 / 0.509 & 0.127 / 0.504 & \underline{0.166 / 0.525} & \textbf{0.178 / 0.532} & \cellcolor{gray!30}7.2 / 1.3 \\
\textit{Sitting Up (a)} 
& 0.149 / 0.507 & 0.140 / 0.511 & 0.171 / 0.517 & 0.163 / 0.515 & \underline{0.187 / 0.537} & \textbf{0.204 / 0.538} & \cellcolor{gray!30}9.1 / 0.2 \\
\textit{Standing up (a)} 
& 0.153 / 0.507 & 0.142 / 0.509 & 0.163 / 0.512 & 0.151 / 0.512 & \underline{0.173 / 0.528} & \textbf{0.191 / 0.535} & \cellcolor{gray!30}10.4 / 1.3 \\
\bottomrule

\end{tabular}}
\label{tab:scripted_vs_baseline}
\vspace{-5pt}
\end{table*}

\begin{table*}[htbp]
\centering
\vspace{-5pt}
\caption{Activity recognition performance (MCC/Macro-F1 scores)  on the \textit{Extrasensory} dataset.}
\vspace{-5pt}
\scalebox{.85}{
\begin{tabular}{|l|c|c|c|c|c|c|c|c|r|}
\toprule
\textbf{Activity} & \textbf{ExtraMLP} & \textbf{LightGBM} & \textbf{CRUFT} & \textbf{HHGNN} & \textbf{DHC-HGL} & \textbf{{\model}} & \textbf{Improv.(\%)} \\ \midrule

\textit{Lying Down}
& 0.936 / 0.968 & 0.911 / 0.955 & 0.968 / 0.984 & 0.962 / 0.981 & \underline{0.977 / 0.989} & \textbf{0.991 / 0.996} & \cellcolor{gray!30}1.4 / \cellcolor{gray!30}0.7 \\
\textit{Sitting} 
& 0.797 / 0.898 & 0.755 / 0.876 & 0.897 / 0.948 & 0.870 / 0.935 & \underline{0.907 / 0.953} & \textbf{0.964 / 0.982} & \cellcolor{gray!30}6.3 / \cellcolor{gray!30}3.0 \\
\textit{Walking}
& 0.554 / 0.739 & 0.560 / 0.746 & 0.709 / 0.838 & 0.718 / 0.846 & \underline{0.772 / 0.878} & \textbf{0.895 / 0.945} & \cellcolor{gray!30}15.9 / \cellcolor{gray!30}7.6 \\
\textit{Sleeping} 
& 0.950 / 0.975 & 0.934 / 0.967 & 0.977 / 0.988 & 0.979 / 0.989 & \underline{0.985 / 0.993} & \textbf{0.994 / 0.997} & \cellcolor{gray!30}0.9 / \cellcolor{gray!30}0.4 \\
\textit{Talking} 
& 0.681 / 0.822 & 0.697 / 0.833 & 0.834 / 0.912 & 0.858 / 0.927 & \underline{0.895 / 0.946} & \textbf{0.945 / 0.972} & \cellcolor{gray!30}5.6 / \cellcolor{gray!30}2.7 \\
\textit{Bath-Shower} 
& 0.496 / 0.695 & 0.729 / 0.848 & 0.654 / 0.801 & 0.780 / 0.880 & \underline{0.827 / 0.908} & \textbf{0.857 / 0.924} & \cellcolor{gray!30}3.6 / \cellcolor{gray!30}1.8 \\
\textit{Toilet} 
& 0.429 / 0.652 & 0.524 / 0.716 & 0.599 / 0.767 & 0.670 / 0.813 & \underline{0.743 / 0.859} & \textbf{0.846 / 0.918} & \cellcolor{gray!30}13.9 / \cellcolor{gray!30}6.9 \\
\textit{Standing} 
& 0.621 / 0.787 & 0.573 / 0.760 & 0.797 / 0.891 & 0.766 / 0.875 & \underline{0.828 / 0.911} & \textbf{0.920 / 0.959} & \cellcolor{gray!30}11.1 / \cellcolor{gray!30}5.3 \\
\textit{Running} 
& 0.616 / 0.774 & 0.656 / 0.802 & 0.733 / 0.850 & 0.866 / 0.930 & \underline{0.883 / 0.938} & \textbf{0.945 / 0.972} & \cellcolor{gray!30}7.0 / \cellcolor{gray!30}3.6 \\
\textit{Stairs-Going Down} 
& 0.441 / 0.661 & 0.791 / 0.886 & 0.618 / 0.780 & 0.695 / 0.831 & \underline{0.807 / 0.898} & \textbf{0.870 / 0.932} & \cellcolor{gray!30}7.8 / \cellcolor{gray!30}3.8 \\
\textit{Stairs-Going Up} 
& 0.446 / 0.664 & 0.654 / 0.801 & 0.651 / 0.801 & 0.706 / 0.837 & \underline{0.770 / 0.876} & \textbf{0.886 / 0.940} & \cellcolor{gray!30}15.1 / \cellcolor{gray!30}7.3 \\
\textit{Exercising} 
& 0.665 / 0.808 & 0.735 / 0.852 & 0.792 / 0.887 & 0.830 / 0.909 & \underline{0.868 / 0.930} & \textbf{0.945 / 0.972} & \cellcolor{gray!30}8.9 / \cellcolor{gray!30}4.5 \\
\bottomrule

\end{tabular}}
\label{tab:extra_vs_baseline}
\vspace{-10pt}
\end{table*}

\subsubsection{Human Activity Recognition Performance}
\label{sec:overallperformance}

As shown in Table~\ref{tab:activity}, {\model} consistently outperformed all baselines across all datasets. It achieved 0.504 / 0.712, 0.921 / 0.959, and 0.922 / 0.959 MCC/Macro-F1 scores on \textit{WASH Scripted}, \textit{WASH Unscripted}, and \textit{Extrasensory} dataset, respectively, thus improving on the performance of the best baselines by 7.8\% / 3.9\% to 22.6\% / 8.4\% on MCC/Macro-F1 scores. The average improvement on activity labels is higher than the average improvement on context labels (see Table~\ref{tab:context}), especially in the WASH datasets. This suggests that activity labels, which are inherently more varied and nuanced, benefit more from LM's added semantic richness. 

The best performing baseline, \textit{DHC-HGL}, initializes context and activity nodes by aggregating handcrafted features associated with each node. As information propagates through the GNN, it learns predictive label representations that capture \underline{co-occurrence correlations} among activities and contexts. On the other hand, our model takes a different approach to capture these label relationships. It introduces semantic label information through LM encoding in an innovative way. As a result, our model consistently outperforms DHC-HGL, the best baseline on all datasets, inspiring confidence in its effectiveness. 
By incorporating semantic label representations, our model excels in recognizing nuanced, confusing, and ill-posed activities even in unconstrained real-world scenarios.

\subsubsection{Detailed Activities}
\label{sec:detailResult}
We provide detailed results for the \textit{WASH Scripted} and \textit{Extrasensory} datasets. Due to its scripted data collection protocol, the \textit{WASH Scripted} dataset contains high-fidelity data. The \textit{Extrasensory} dataset is publicly available, enabling other researchers to attempt reproducing our results and reflecting more complex, real-world, unscripted scenarios. Due to space limitations, the {\model} model's performance on \textit{WASH Unscripted} was omitted,  while it also follows a consistent pattern, with similar observations and conclusions. 

Overall, {\model} performed well on activities such as \textit{Talking (On Phone)}, \textit{Bathroom(-Shower)}, and \textit{Toilet}, which could be attributed to those activities' more distinct patterns and precise semantic meanings that our model can capture. The controlled lab environment in the \textit{WASH Scripted} data collection allowed the inclusion of some rare activities, such as \textit{Trembling}, \textit{Coughing}, and \textit{Sneezing}, along with short-term actions such as \textit{Laying Down (action)}, \textit{Sitting Down (action)}, \textit{Sitting Up (action)}, and \textit{Standing Up (action)}. {\model} exhibited exceptional recognition of these activities, achieving average MCC/Macro-F1 improvements of 26.3\% / 8.4\% on rare activities and 8.7\% / 1.1\% on short-term actions. For activities such as \textit{Trembling}, \textit{Coughing}, and \textit{Sneezing}, sensor signals may not exhibit significant changes, especially in the absence of audio sensors. The semantic information from encoding labels helped the model detect these activities reliably. Similarly, for short-term actions, the label sentences indicated their brief nature, guiding the model to concentrate on the abrupt changes in sensor signals. This approach allowed the {\model} to more accurately identify the activities based on the enriched label descriptions, even when the raw sensor data was less distinctive.

\begin{table*}[htbp]
\centering
\vspace{-10pt}
\caption{Average context recognition performance on the \textit{WASH Scripted}, \textit{WASH Unscripted}, and \textit{Extrasensory} datasets.}
\vspace{-5pt}
\scalebox{.85}{
\begin{tabular}{|l|c|c|c|c|c|c|c|c|r|}
\toprule
\textbf{Dataset} & \textbf{ExtraMLP} & \textbf{LightGBM} & \textbf{CRUFT} & \textbf{HHGNN} & \textbf{DHC-HGL} & \textbf{{\model}} & \textbf{Improv.(\%)} \\ \midrule
\textit{WASH Scripted} & 0.719 / 0.845 & 0.705 / 0.839 & 0.707 / 0.836 & 0.765 / 0.875 & \textbf{0.807 / 0.897} & \underline{0.796 / 0.890} & -1.4 / -0.8 \\ \midrule
\textit{WASH Unscripted} & 0.686 / 0.823 & 0.687 / 0.827 & 0.704 / 0.835 & 0.813 / 0.901 & \underline{0.902 / 0.950} & \textbf{0.960 / 0.980} & \cellcolor{gray!30}6.4 / \cellcolor{gray!30}3.2 \\\midrule
\textit{Extrasensory} & {0.784 / 0.883} & {0.835 / 0.913} & {0.897 / 0.946} & \underline{0.953 / 0.976} & {0.950 / 0.975} & \textbf{0.977 / 0.988} & \cellcolor{gray!30}{2.5 / \cellcolor{gray!30}1.2} \\\bottomrule
\end{tabular}}
\label{tab:context}
\vspace{-5pt}
\end{table*}

\begin{figure*}[ht]
    \centering
    \includegraphics[width=0.3\linewidth]{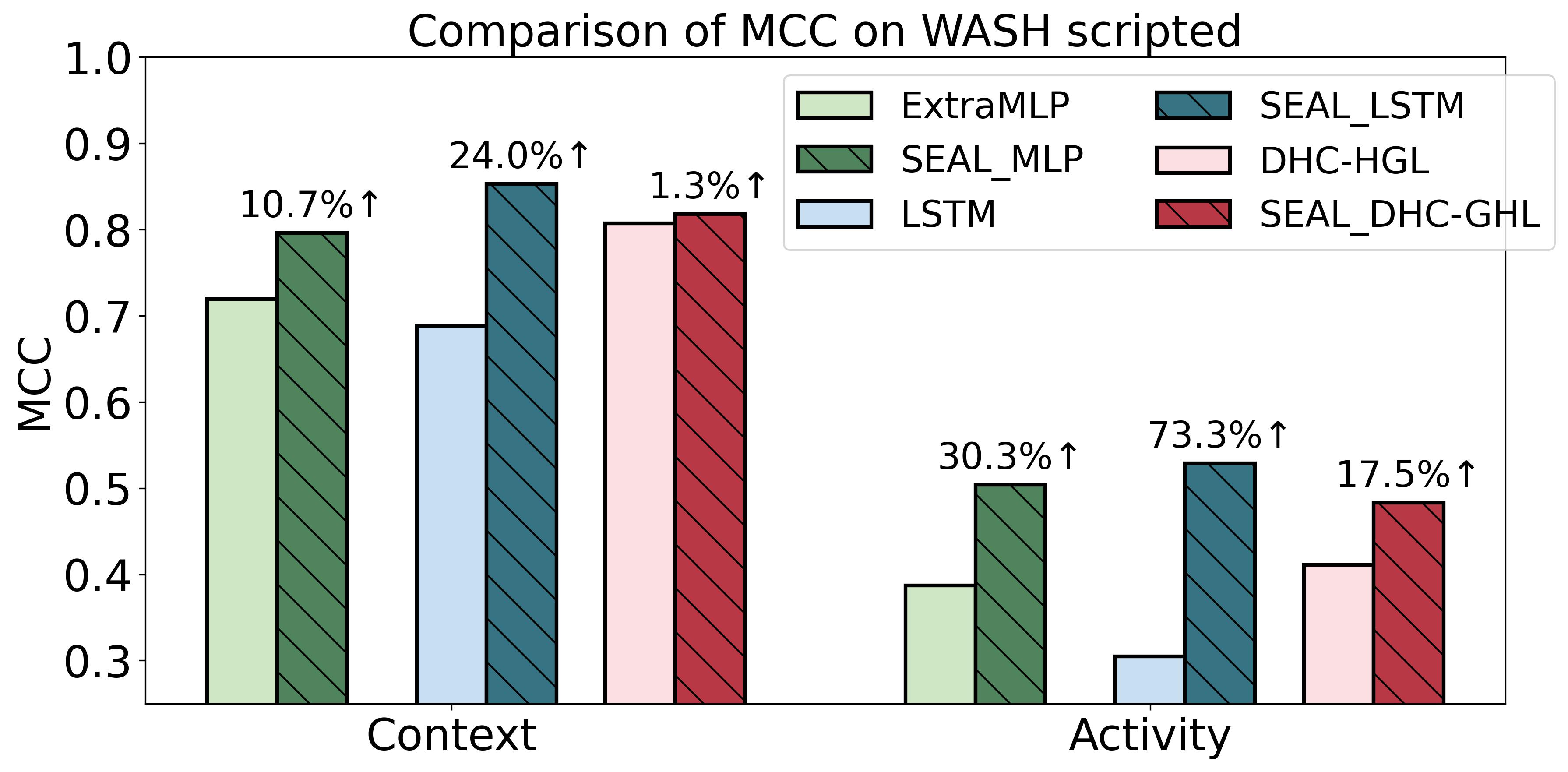}
    \includegraphics[width=0.3\linewidth]{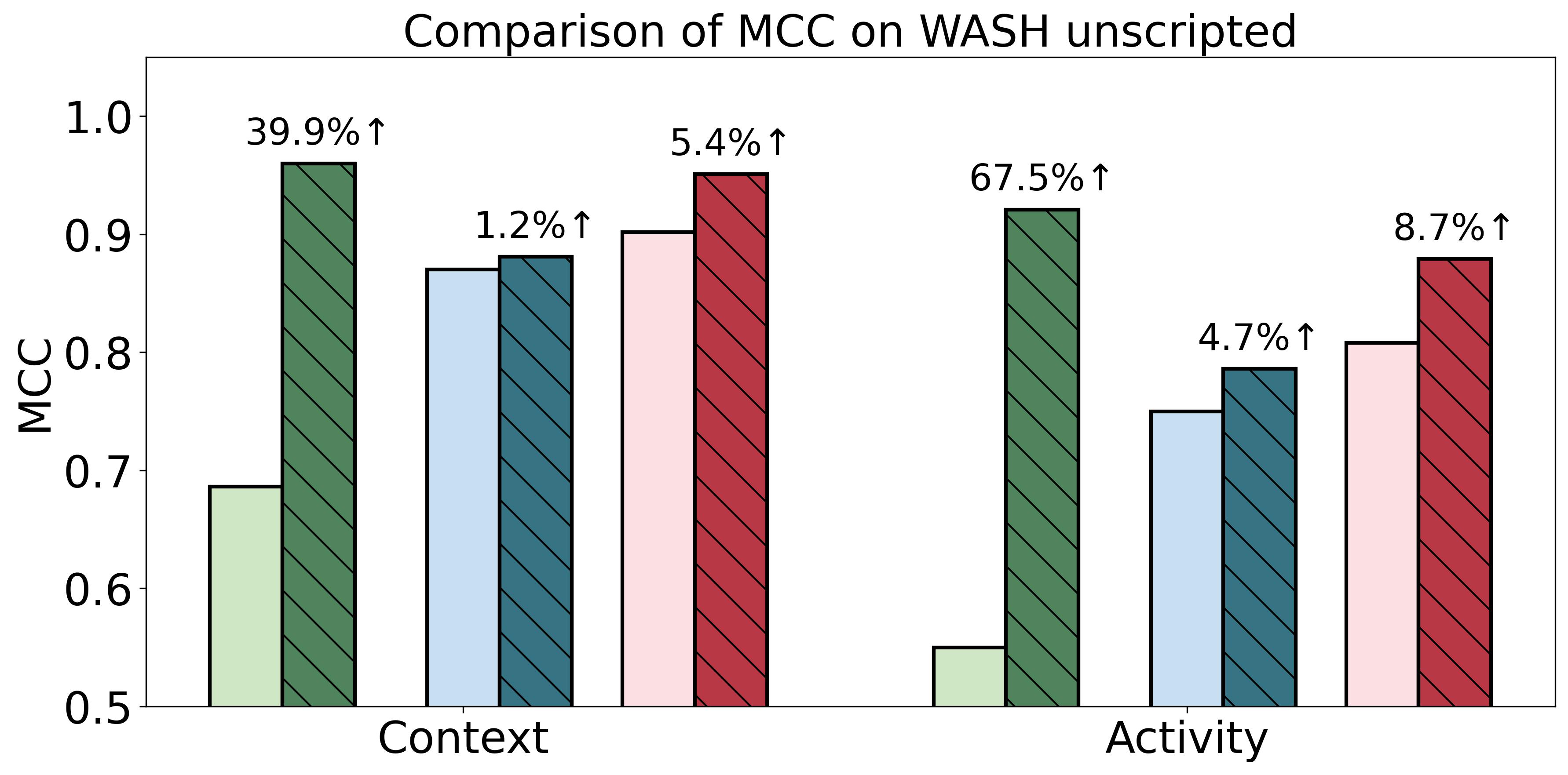}
    \includegraphics[width=0.3\linewidth]{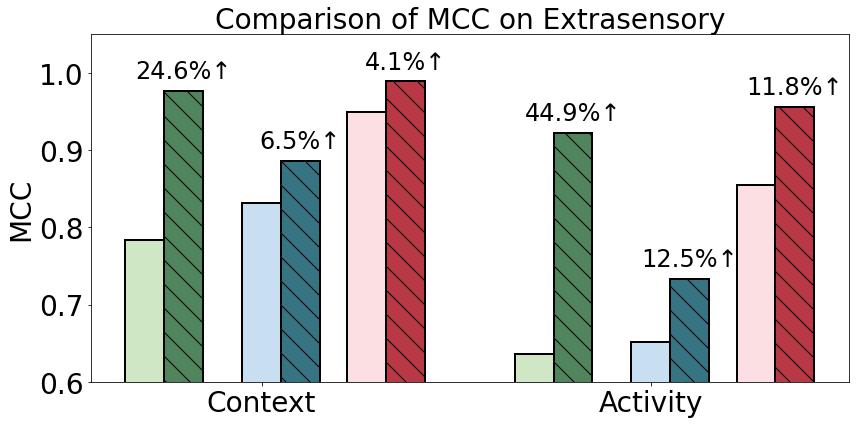}

    
    \includegraphics[width=0.3\linewidth]{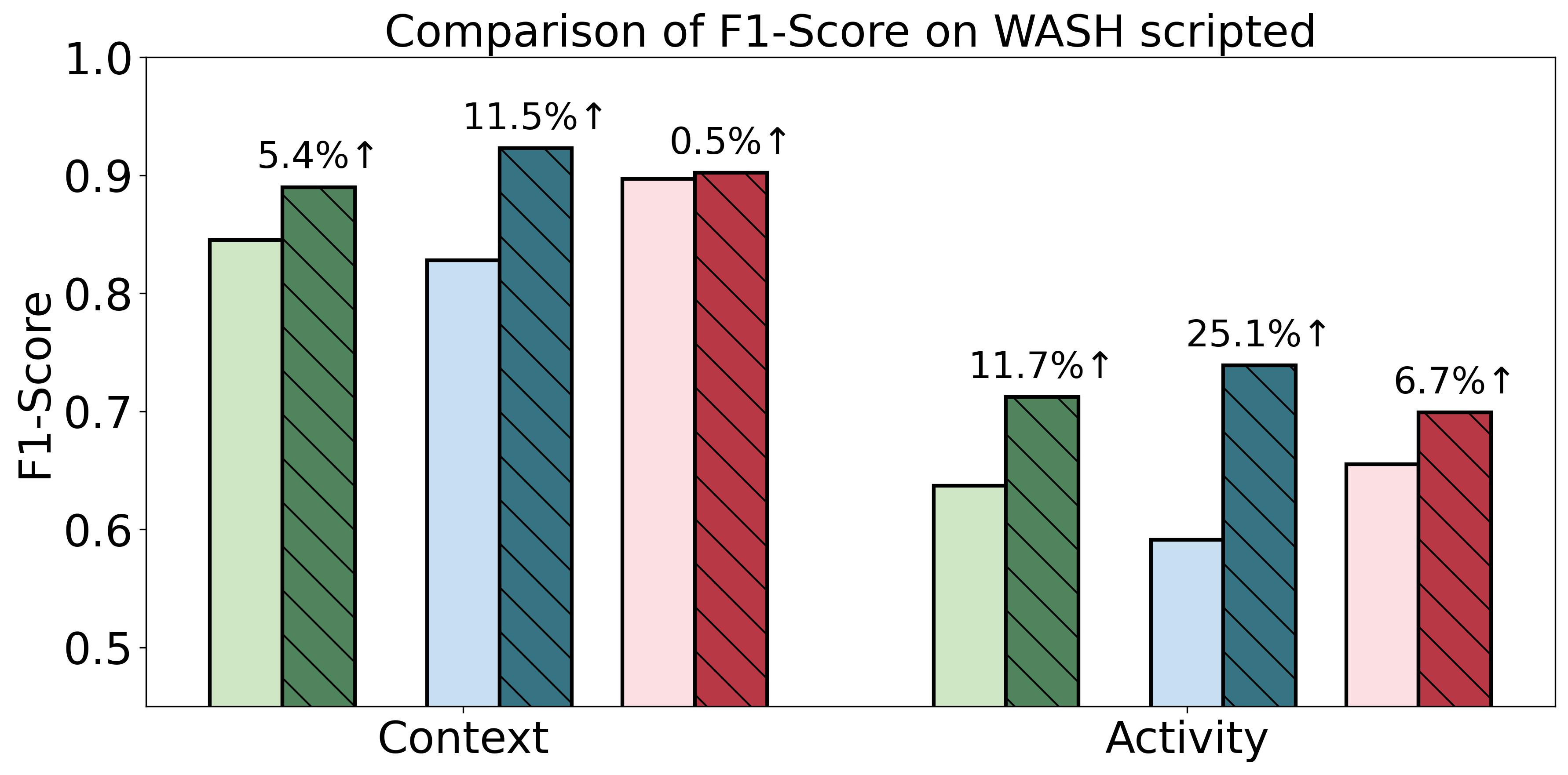}
    \includegraphics[width=0.3\linewidth]{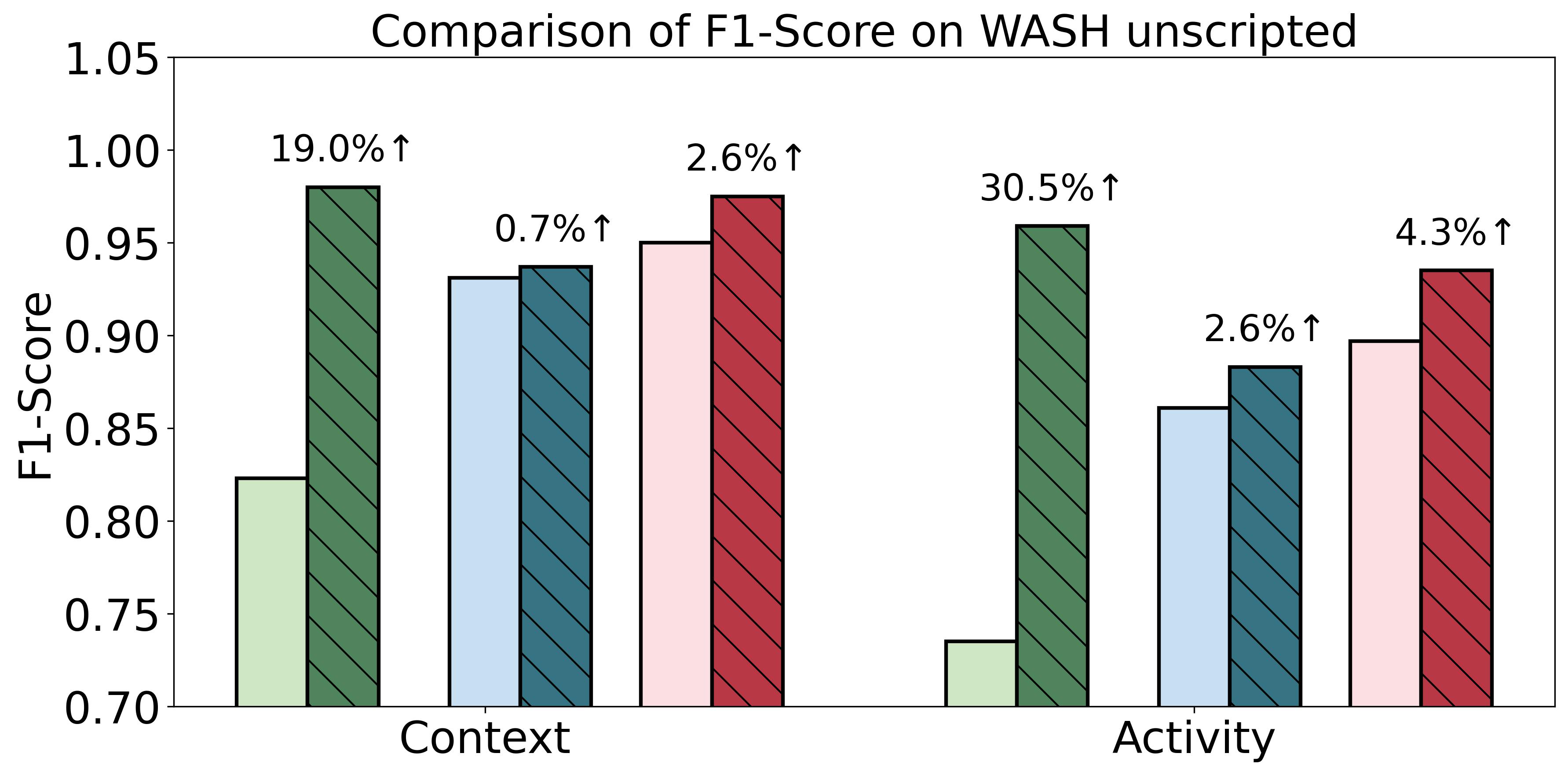}
    \includegraphics[width=0.3\linewidth]{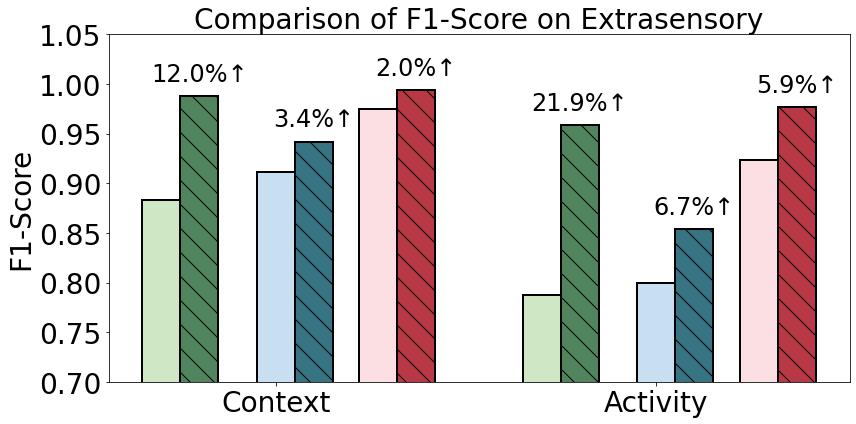}
    \vspace{-5pt}
    \caption{Result of {\model} using different backbones. We observe that {\model} can show improvement with all backbones across all datasets.}
    \label{fig:backbones}
    \vspace{-10pt}
\end{figure*}

For labels that are often misclassified between each other, such as \textit{Jogging} vs. \textit{Running}~\cite{chakraborty2008view} and \textit{Stairs Going Up} vs. \textit{Stairs Going Down}~\cite{fridriksdottir2020accelerometer}, {\model} achieved even more noticeable improvements. Additionally, we observed impressive improvements on activities with hierarchical structures, such as \textit{Exercising} (which encompasses \textit{Walking}, \textit{Running}, \textit{Jogging}, and \textit{Jumping}). SEAL delivered outstanding recognition performance, particularly in distinguishing activities that other CA-HAR models have historically confused (e.g., differentiating between Going Up and Going Down Stairs). Furthermore, it performed remarkably well on activities that are difficult to classify based solely on sensor data, such as \textit{Talking On Phone}, where audio data is ideal for accurate recognition.

\vspace{-5pt}
\section{Analysis}
\label{sec:analysis}
\vspace{-5pt}
\subsection{{\model}'s performance on context recognition}
The average context recognition performance across all datasets is listed in Table~\ref{tab:context}. {\model} achieved MCC/Macro-F1 scores of 0.796/0.890, 0.960/0.980, and 0.977/0.988 on the three datasets, respectively. Notably, {\model} significantly improved context recognition performance on the two unscripted datasets, where how users held their phones while performing various activities was not constrained. This reflects the model's strength in handling complex, real-world scenarios with more diverse and unpredictable conditions.
The performance dropped slightly for context recognition in \textit{WASH Scripted}. This might be due to the nature of this dataset, where activities and contexts were performed under specific instructions, leading to less natural variability. Additionally, as pointed out in the dataset section, the \textit{WASH Scripted} dataset has fewer instances, which could make it more susceptible to noise, causing the {\model} to misinterpret the context.

\subsection{{\model} with different backbones}
We hypothesize that semantic encoding could improve activity recognition in two ways: 1) directly for inference and 2) as a source of supplementary information to enhance other label encoding methods, such as Graph Neural Networks (GNNs). To evaluate the merits of these two semantic encoding approaches, we explored three types of deep learning backbones, spanning models that only leveraged handcrafted features or raw signals to models that obtained label representation using GNNs: \textit{ExtraMLP}, \textit{LSTM}, and \textit{DHC-HGL}. The results are shown in Fig.~\ref{fig:backbones} where LM label encoding provides backbone-agnostic improvements with consistent improvements in both the MCC and Macro-F1 scores across all datasets. Notably, {\model} achieved the most significant improvements on \textit{ExtraMLP} and \textit{LSTM}. Their MCC / Macro-F1 scores increased from 1.2\% / 0.7\% to 39.9\% / 19.0\% for context labels and from 4.7\% / 2.6\% to 73.3\% / 30.5\% for activity labels.

Even though DHC-HGL inherently captures label relationships, it still benefits from the language model's semantic encoding, achieving MCC/Macro-F1 improvements ranging from 1.3\% / 0.5\% to 5.4\% / 2.6\% for context labels and from 8.7\% / 4.3\% to 17.5\% / 6.7\% for activity labels. Although the degree of improvement for DHC-HGL is less than that of \textit{ExtraMLP} and \textit{LSTM}, language modeling can further enrich label information even for models already adept at capturing inherent relationships.
Additionally, while context and activity labels exhibited gains from incorporating the Language Model encodings, the improvements were more substantial for activity labels across all datasets. This suggests that {\model} is particularly effective in making accurate predictions in scenarios with varied activity patterns, which are likely more intricate than contextual information. Interestingly, {\model} outperformed {\model}\_DHC-HGL on the two WASH datasets (Scripted and Unscripted), further validating the effectiveness of leveraging the Language Model's semantic capabilities.

Aside from these analyses, we also provide visualizations of the learned label embeddings across all datasets along with an in-depth analysis in the appendix. Please refer to Sec.~\ref{sec:visualization}.

\section{Conclusion}
In this paper, we proposed {\model}, a novel framework that leverages language models for CA-HAR. Unlike traditional HAR models that overlook semantic relationships among labels, {\model} directly integrates LMs into the decision-making process, capturing the semantic relationships for improved activity recognition. Experimental results demonstrated that {\model} outperforms SOTA CA-HAR models, achieving 0.504/0.712, 0.921/0.959, and 0.922/0.959 MCC/Macro-F1 scores across three datasets. The results further highlight its ability to distinguish semantically similar activities with subtle differences or abrupt changes. These findings demonstrate the effectiveness of incorporating LMs into CA-HAR tasks and suggest promising directions for extending this approach to multi-modal data incorporating additional modalities, including image and hierarchical label structure.

\section*{Acknowledgment}
This research is supported by NSF grant IOS-2430277 and DARPA grant HR00111780032-WASH-FP-031. 

\bibliographystyle{IEEEtran}
\bibliography{refs}

\pagebreak

\begin{appendices}
\begin{table*}[ht]
    \centering
    \caption{Comparison of 3 benchmark Context-Aware Human Activity Recognition datasets. Accelerometer, gyroscope, magnetometer, and gravity are tri-axial sensors. The sampling rate (Hz) is listed in parentheses. If no sampling rate was specified, the sensor was sampled once per minute or when its value changed more than a threshold. (a) stands for short-term actions.}
    \vspace{-5pt}
    \scalebox{.96}{
    \begin{tabular}{l|c|c|c|c}
    \toprule  
    \textbf{Information} & \textbf{Dataset} & \textbf{WASH Scripted} & \textbf{WASH Unscripted} & \textbf{Extrasensory (Unscripted)} \\
    \midrule
    \midrule
    \multirow{1}{*}{General} & Instance/Feature/User & 294,512 / 144 / 107 & 7,773,479 / 139 / 108 & 6,355,350 / 170 / 60 \\
    \midrule
    \midrule
    \multirow{3}{*}{Sensor} & Unique Sensors & Calibrated Magnetometer(40) & Calibrated Magnetometer(40) & Calibrated Gravity(40), Audio(46) \\ \cmidrule{2-5}
    &\multirow{2}{*}{Common Sensors}
    & \multicolumn{3}{c}{\shortstack{Raw Accelerometer(40), Calibrated Gyroscope(40), Phone State}} \\
    && \multicolumn{3}{c}{\shortstack{Raw Magnetometer(40), Location, Environment Measure}} \\
    \midrule
    \midrule
    \multirow{9}{*}{Label} & Number of Context & 5 & 5 & 4 \\
    & Unique Context & On Table & On Table & On Table-(Face Down, Face Up) \\ \cmidrule{2-5}
    & Common Context & \multicolumn{3}{c}{In Pocket, In Hand, In Bag} \\
    \cmidrule{2-5}
    &Number of Activity & 20 & 13 & 12 \\
    &\multirow{4}{*}{Unique Activity} 
    & Talking on Phone, Bathroom, Jogging & Talking on Phone & Talking\\ 
    && Typing, Coughing, Jumping, Sneezing & Exercising & Bath-Shower \\ 
    && Trembling, Laying Down(a), Sitting Down(a) & Bathroom & Toilet\\ 
    && Sitting Up(a), Standing Up(a) & Jogging, Typing & Exercising \\ \cmidrule{2-5}
    &Common Activity & \multicolumn{3}{c}{\shortstack{Lying Down, Sitting, Walking, Sleeping, Standing, Running, Stairs-Going Down, Stairs-Going Up}} \\

    \bottomrule
    \end{tabular}}
    \label{tab:datasets}
\end{table*}

\section{Datasets}
\label{sec:dataset}
We rigorously evaluated {\model}'s performance on three CA-HAR datasets, spanning both scripted and unscripted data collection methods. The scripted dataset, \textit{WASH Scripted} ({\url{https://tinyurl.com/darpaWash}}), involves participants performing activities in a controlled, pre-defined sequence, resulting in a smaller but more reliable dataset. In contrast, the unscripted datasets, \textit{WASH Unscripted} and \textit{Extrasensory}~\cite{vaizman2017recognizing}, were collected as participants went about their daily lives without any constraints, periodically self-reporting context labels. Unscripted datasets reflect natural human behavior better and are typically larger than scripted datasets since data collection is done continuously as the subject lives their lives; however, they may contain noisy, incomplete, or wrong data due to memory decay or recall bias when subjects report labels retrospectively. A detailed summary of the statistics for all three CA-HAR datasets can be found in Table~\ref{tab:datasets}. Although context can refer to any external information (e.g., location, time, user profile, device placement, and previous activities that improves understanding of human activities, in this work, we focus on phone placement. We define context-aware human activity the as $<$activity, phone placement$>$ tuple.

\subsection{Data Preprocessing}
To ensure data quality and consistency in the unscripted dataset, where incorrect labels might be present, instances with conflicting labels that could not feasibly co-occur were removed from the datasets.
For example, samples labeled with contradicting phone placements (e.g., \textit{Phone on Table} and \textit{Phone in Pocket}) or incompatible activities (e.g., \textit{Sleeping} while \textit{Running}) were excluded. After filtering, informed by parameters prior work found to be optimal~\cite{ge2020cruft}, sensor data was segmented using a sliding overlapping window approach, with a window size of 3 seconds and a step size of 1.5 seconds, generating data instances for further analysis. 

\subsection{Feature Extraction}
For CA-HAR models that automatically learn features from raw sensor signals~\cite{ge2020cruft,li2019bi}, we resampled each segmented 3-axial sensor signal instance into a fixed-length sequence using the Fourier method~\cite{cooley1965algorithm} (Eq.~\ref{eq:fourier}), resulting in a consistent representation of $x^{'}_{raw} \in \mathcal{N}^{12\times50}$ samples per window, where $i=\sqrt{-1}$ is the imaginary unit, $T$ denotes the original signal length and $K$ is the resampled signal length. This step ensures uniform length across all instances, making them suitable for deep-learning models.
\begin{equation}
\scriptsize
\begin{aligned}
   & X_{raw} = [x^i_T]_{i=1}^{n} \Rightarrow X^{'}_{raw} = [x^i_K]_{i=1}^{n}\\
   & x_k = \sum_{t=0}^T x_t e^{-i2 \pi kt/T}, k=1, ..., K 
\end{aligned}
    \label{eq:fourier}
\end{equation}

For CA-HAR models that analyze handcrafted features~\cite{vaizman2018context, ge2020cruft,ke2017lightgbm,ge2023heterogeneous,ge2022qcruft}, we extracted predictive features for CA-HAR tasks, as recommended by previous research~\cite{vaizman2018context,ge2020cruft,ge2022qcruft,ge2023heterogeneous,ge2024deep}. 
Features with constant values across all instances in the training set were removed during preprocessing, as they were not discriminative between different classes. After this filtering process, 144, 139, and 170 features were retained for the three datasets, respectively. 
Data from each user was split at the instance level into training (60\%), validation (20\%), and testing (20\%) sets. 
We then normalized the features by subtracting the mean and scaling to unit variance $x = (f-\mu)/s$ based on the training set, and applied the same normalization to the validation and test sets. All missing values were filled with zeros.

%

\section{Experiment Settings}
\label{sec:expSetting}

Bayesian optimization with a Gaussian process~\cite{gardner2014bayesian} was used to determine optimal hyperparameter values for all models with more details in Sec.~\ref{sec:hyper}. Additionally, the RAdam~\cite{liu2019variance} optimizer and an exponential learning rate scheduler were used.

\subsection{Hyperparameters}

The optimal hyperparameters for our models were selected based on the validation loss to ensure robust generalization. A comprehensive search was conducted within a pre-defined range using Bayesian optimization. The specific search ranges and the final selected values for each hyperparameter are summarized in Table~\ref{tab:hypers}.

\label{sec:hyper}
\begin{table*}[ht]
    \centering
    \caption{Hyperparameters, Search Space, and Selected Values for {\model} Training.}
    \vspace{-5pt}
    \begin{tabular}{|r|l|c|c|c|c|}
    \toprule
        \textbf{Hyper} & \textbf{Description} & \textbf{Search Space} & \textit{WASH Scripted} & \textit{WASH Unscripted} & \textit{Extrasensory} \\\midrule
        lr  & Learning rate & [$10^{-7}, 10^{-3}]$ & $10^{-3.44}$ &  $10^{-4.27}$ &  $10^{-4.40}$ \\
        \midrule
        epoch & Epoch & [100, 800] & 151 & 499 & 713\\\midrule
        $h_2'$ & Dimension of 1st hidden layer& [256, 2048] &537 & 1583 & 824 \\\midrule
        $h_2$ & Dimension of 2nd hidden layer& [256, 2048] & 1005 & 752 & 376 \\\midrule
        $h$ & Dimension of common vector space& [256, 4096] & 654 & 1871 & 3683\\\midrule
        dropout & Dropout between hidden layers&[0, 0.5] & 0.21& 0.02 & 0.02\\\bottomrule
    \end{tabular}
    \label{tab:hypers}
\end{table*}

\subsection{Evaluation Metrics}
\label{sec:metrics}
 Matthews Correlation Coefficient (MCC) and Macro F1 score, which are well-suited for evaluating for multi-label classification tasks on imbalanced datasets, were used to evaluate model performance. MCC measures  classification performance while considering T/F P/N (Eq.~\ref{eq:mcc}) on imbalanced datasets. The MCC ranges from -1 (completely incorrect predictions) to +1 (perfect predictions).
\begin{equation}
    \scriptsize
    MCC = \frac{TP \times TN - FP \times FN}{\sqrt{(TP+FP)(TP+FN)(TN+FP)(TN+FN)}}
    \label{eq:mcc}
\end{equation}
The F1 score for each label is calculated as in E.q.~\ref{eq:f1}, where $Precision = \frac{TP}{TP+FP}$ and $Recall = \frac{TP}{TP+FN}$. The Macro F1 scores (Eq.~\ref{eq:macrof1}) evaluate how well the model performed across all labels by averaging the F1 score across all labels. The Macro F1 score treats each label equally, ensuring that minority classes are appropriately evaluated.
\begin{equation}
\scriptsize
    F1 = 2 \times \frac{Precision \times Recall}{Precision + Recall}
    \label{eq:f1}
    \vspace{-5pt}
\end{equation}
\begin{equation}
\scriptsize
    Macro\ F1 = \frac{1}{C}\sum_{i=1}^C F1_i
    \label{eq:macrof1}
\end{equation}

\section{Baselines}
\label{sec:baseline}
\smallskip\noindent \textbf{ExtraMLP~\cite{vaizman2018context}:} A Multi-Layer Perceptron (MLP) neural networks method proposed initially for CA-HAR on the Extrasensory dataset. Using handcrafted features, ExtraMLP recognized contexts and activities. Custom weights were used to handle imbalanced, incomplete data.
    
\smallskip\noindent\textbf{LightGBM~\cite{ke2017lightgbm}:} An algorithm widely utilized in academia and industry. We trained separate models for each label so that each model could focus exclusively on learning patterns related to a single label, reducing interference between similar/easily confused labels.

\smallskip\noindent \textbf{CRUFT~\cite{ge2020cruft}:} A state-of-the-art CA-HAR method leveraging handcrafted features and raw signal inputs on two different MLP and CNN-Bi-LSTM branches, respectively. It captured the temporal relationships between instances and estimated prediction uncertainty to provide reliable predictions.

\smallskip\noindent \textbf{HHGNN~\cite{ge2023heterogeneous}:} A state-of-the-art graph learning CA-HAR method leveraging a heterogeneous hypergraph neural network to encode the complex relationship among users, phone placements, and activity and transformed the classification problem into node presentation learning problem. 

\smallskip\noindent \textbf{DHC-HGL~\cite{ge2024deep}:} A state-of-the-art graph learning CA-HAR method that employs contrastive loss and constructs specific sub-hypergraphs. It enhanced the model's ability to learn representative label encodings, thereby facilitating more effective CA-HAR. HHGNN and DHC-HGL were included as baselines to assess whether language modeling captures context-activity relationships better and delivered more predictive label representations than graph-based approaches.

\section{Visualization of learned label embedding}
\label{sec:visualization}
\begin{figure*}[t]
    \centering
    \includegraphics[width=0.32\linewidth]{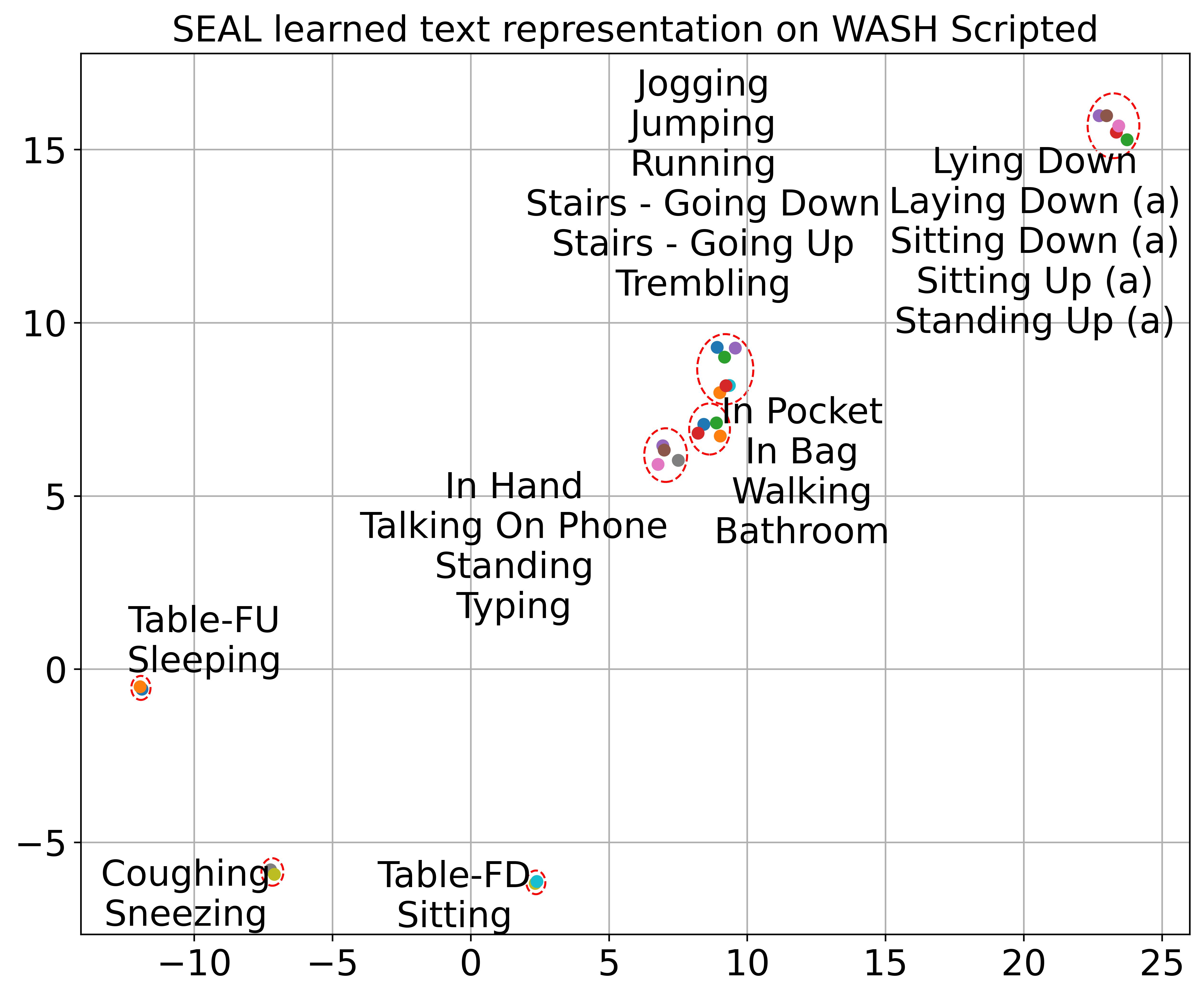}
    \includegraphics[width=0.33\linewidth]{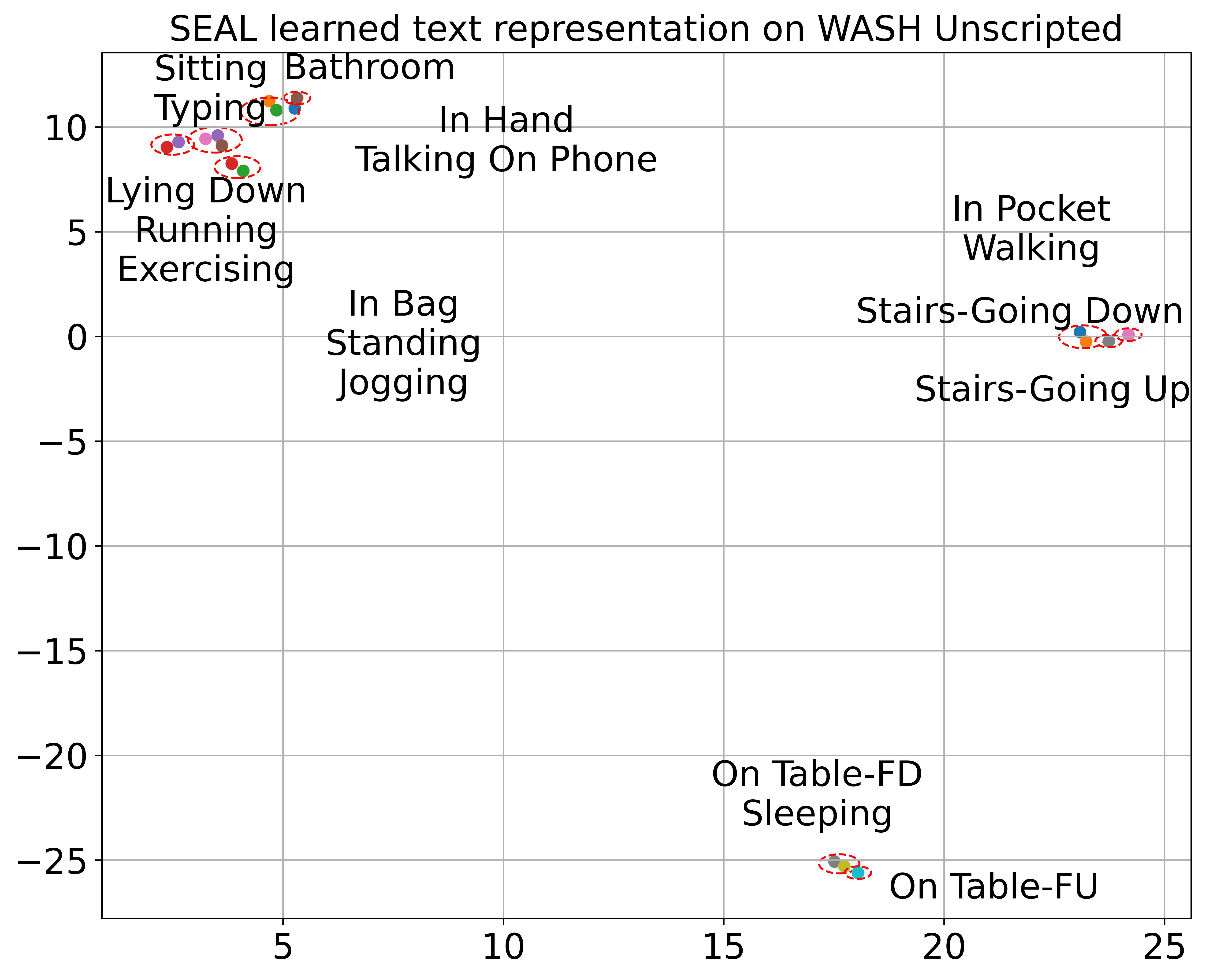}
    \includegraphics[width=0.32\linewidth]{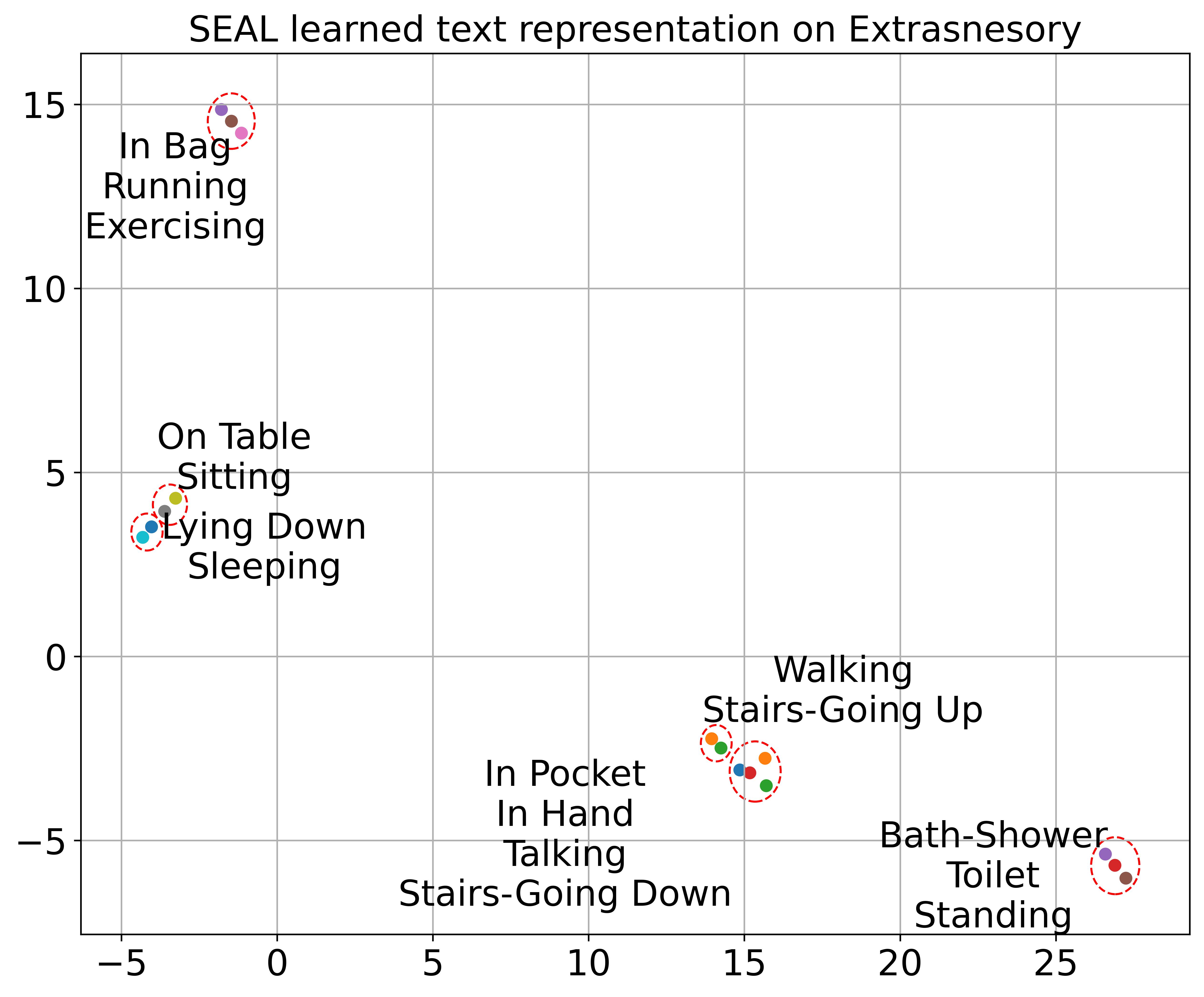}
    \vspace{-5pt}
    \caption{UMAP visualization of {\model} learned text embedding across all datasets, with clusters generated by KMeans. The clusters indicate {\model}'s ability to capture the semantic relationship between activities and contexts.}
    \label{fig:umap}
    \vspace{-5pt}
\end{figure*}

We hypothesize that by leveraging Language Model (LM) to encode CA-HAR labels, {\model} can capture the semantic information of labels and the inherent relationships between context and activities. To validate this hypothesis, in Fig.~\ref{fig:umap}, we present UMAP~\cite{mcinnes2018umap} visualizations of the learned text representations generated by {\model} across all datasets, with clustering using KMeans~\cite{lloyd1982least}. We can observe that \textit{Stairs-Going Up} and \textit{Stairs-Going Down}; \textit{Exercising} and \textit{Running}; \textit{Toilet} and \textit{Bathroom-Shower} lie close to each other, reflecting {\model}'s ability to correctly encode the semantic information of activities. Additionally, some clusters were formed due to the context and activity's inherent relationships, and {\model} learned such relationships from co-occurrence in data rather than just semantic similarities. For example, \textit{On Table(-FD/U)} and \textit{Sleeping} are clustered close together, since people usually put their phones down when sleeping. Similarly, \textit{Talking (On Phone)} and \textit{In Hand} are also clustered closely, which is intuitive because people generally hold their phones while talking. Moreover, in the \textit{WASH Scripted} dataset, \textit{Coughing} and \textit{Sneezing} formed a cluster, and all short-term actions lie close to each other, also serving as evidence that our model's strength in semantic encoding. These observations are consistent with our quantitative analyses in Sec.~\ref{sec:detailResult}.

\end{appendices}

\end{document}